\documentclass{article} 
\usepackage{iclr2026_delta,times}

\usepackage{amsmath,amsfonts,bm}

\def\eqref#1{equation~\ref{#1}}

\def\1{\bm{1}}

\DeclareMathAlphabet{\mathsfit}{\encodingdefault}{\sfdefault}{m}{sl}
\SetMathAlphabet{\mathsfit}{bold}{\encodingdefault}{\sfdefault}{bx}{n}

\usepackage{hyperref}
\usepackage{url}
\usepackage[T1]{fontenc}    
\usepackage{booktabs}
\usepackage{amsfonts}
\usepackage{nicefrac}
\usepackage{microtype}      
\usepackage[table, dvipsnames]{xcolor}
\usepackage{amsmath}
\usepackage{float}
\usepackage[ruled,vlined,linesnumbered]{algorithm2e} 
\usepackage{multirow, makecell}
\usepackage{graphicx}
\usepackage{subcaption}
\usepackage{wrapfig}
\usepackage{multirow}
\usepackage{algorithm2e}   
\usepackage{float}         
\usepackage{cuted}         
\iclrfinalcopy

\definecolor{myrowcolour}{rgb}{0.859,0.859,0.859}

\def\meth{MixFlow}
\def\FC{$\kappa\texttt{-FC}$}

\title{\meth{}: Mixed Source Distributions Improve Rectified Flows}

\author{Nazir Nayal  \quad Christopher Wewer \quad Jan Eric Lenssen  \\
Max Planck Institute for Informatics, Saarland Informatics Campus, Germany\\
\texttt{\{nnayal,cwewer,jlenssen\}@mpi-inf.mpg.de}
}

\begin{document}

\maketitle

\begin{abstract}
    Diffusion models and their variations, such as rectified flows, generate diverse and high-quality images, but they are still hindered by slow iterative sampling caused by the highly curved generative paths they learn. An important cause of high curvature, as shown by previous work, is independence between the source distribution (standard Gaussian) and the data distribution. In this work, we tackle this limitation by two complementary contributions. First, we attempt to break away from the standard Gaussian assumption by introducing \FC{}, a general formulation that conditions the source distribution on an arbitrary signal $\kappa$ that aligns it better with the data distribution. Then, we present \meth{}, a simple but effective training strategy that reduces the generative path curvatures and considerably improves sampling efficiency. \meth{} trains a flow model on linear mixtures of a fixed unconditional distribution and a \FC{}-based distribution. This simple mixture improves the alignment between the source and data, provides better generation quality with less required sampling steps, and accelerates the training convergence considerably. On average, our training procedure improves the generation quality by 12\% in FID compared to standard rectified flow and 7\% compared to previous baselines under a fixed sampling budget. Code available at: \href{https://github.com/NazirNayal8/MixFlow}{https://github.com/NazirNayal8/MixFlow}
\end{abstract}


\section{Introduction}

Generative modeling, the problem of fitting and sampling from data distributions, is a heavily explored topic with remarkable success in recent years, mostly driven by progress in image generation~\cite{ddpm, ddim, flow-matching}. Existing generative models offer trade-offs between sampling speed, diversity, and the quality of the generated samples, referred to as the generative learning trilemma \cite{trilemma}. Diffusion models \cite{score-based, ddpm, ddim} and their variations have pushed the performance considerably in terms of diversity and quality. However, a single inference requires several forward passes to obtain high quality samples. Therefore, several works have explored ways to reduce the number of function evaluations required for sampling. Rectified Flow \cite{rectified-flow} and Flow Matching \cite{flow-matching} tackled the problem from the perspective of straightening the generative paths by replacing the diffusion schedulers with optimal transport displacement interpolations \cite{ot_displacement} between the source and target distributions. Even though their formulations provide theoretical guarantees for requiring a fewer number of sampling steps, the amount of required steps still remains high in practice. In this work, we tackle this problem by introducing an effective training strategy for flow models to reduce the amount of steps required to generate high quality samples.

Flow models learn to iteratively transform a simple source distribution, usually a standard Gaussian, to a complex data distribution. For Flow Matching, a recent line of work shows that the sampling speed is strongly influenced by the assumptions on the forward coupling \cite{minibatch-ot, multisample-fm, fast-ode}. The forward coupling is the joint distribution of the source and the target, which encodes their dependence relation.  Optimizing the forward coupling leads to straighter generative paths by exposing the model to source-target pairs that are more aligned \cite{fast-ode}.

Inspired by this, we propose $\kappa$-Forward Coupling (\FC{}), a general formulation for learnable forward couplings that can utilize an arbitrary guiding signal $\kappa$ to align the source distribution with the target. The more informative $\kappa$ is of the data distribution, better alignment is achieved. Nevertheless, we show that naively optimizing the forward coupling with $\kappa$ introduces a difficult trade-off with a regularization hyperparameter that can lead to issues like the prior hole problem \cite{cvae}. To counter this, we introduce \meth{}, a technique that uses a linear mixture of two distributions as the source distribution, one of which is fixed and the other a learnable distribution that is learned using \FC{}. The mixing encourages that samples on the interpolation path map to similar regions in the target distribution, transporting structure from the conditional to the unconditional (Gaussian) source distribution.
\meth{} demonstrates overall improvement in sampling quality and requires fewer sampling steps. Furthermore, we show that, given a conditioning signal that is sufficiently informative, our formulation allows for controlling the speed-quality trade-off at test-time. 

To verify our findings, we present exhaustive results on common image generation benchmarks, showing that our approach improves FID by 12\% compared to standard Rectified Flow and by 7\% compared to the best previous method for trajectory straightening, with a comparable number of sampling steps. In contrast to previous works, our trade-off does not depend on parameters that need to be set during training. We provide an analysis of different design choices and their effect on the source distribution, generation quality, and sampling speed.

In summary, our contributions are:
\begin{itemize}
    \item We propose \FC{}, a general formulation for learnable forward couplings that can be conditioned on arbitrary variables for obtaining better source distributions.
    \item We introduce \meth{}, a method for training Rectified Flows with a linear mixture of two distributions as a source distribution, which leads to less sampling steps required to generate high quality samples.
\end{itemize}

\section{Related work}

In general, the lines of work that study the sampling speed problem in diffusion models can be categorized into the following groups, depending on which part of the design space is examined.

\paragraph{Distillation.} One direction explores linearizing the mapping between the source and the target distribution through distillation~\cite{distillation_salimans2022progressive, rectified-flow, distillation_Berthelot2023TRACTDD, distillation_NEURIPS2024_47ee3941, distillation_Luhman2021KnowledgeDI, distillation_Xie2024EMDF}, consistency constraints~\cite{Song2023ConsistencyM, ict_consistency, vct, ect, consistency_fm}, or Reflow \cite{rectified-flow}. While these methods are able to achieve reasonable generation quality with a single sampling step, they require retraining a model multiple times, and often degrade the model's performance for higher number of sampling steps \cite{variational_rf_matching}. On the other hand, we show that training \meth{} once can improve the performance for all choices of sampling steps, and can also considerably reduce the required training budget. A branch in this direction attempts to improve the Reflow operation by generalizing it to arbitrary schedules \cite{rectified_diffusion}, or enhancing its design components \cite{simple_reflow}. Furthermore, we highlight that this direction is orthogonal to our approach and these methods can be applied to any model trained with \meth{}.

\paragraph{Faster Solvers.} Another line of work focuses on developing faster samplers by utilizing better numerical ODE solvers \cite{samplers_dockhorn2022genie, samplers_NEURIPS2022_a98846e9, samplers_lu2022dpmsolver, ddim}. Despite these improvements, the sampling speed remains bounded by the curvature of the generative trajectories induced by the flow models. In this work, we tackle the same problem but from the orthogonal perspective of source distribution optimization. Hence, our method can be combined with any ODE solver to achieve faster sampling.

\paragraph{Path Straightness.} Rectified Flow \cite{fast-ode} shows that the intersections of the paths constructed by the source and target distribution samples affect the straightness of the generative paths. Flow-Matching \cite{flow-matching} defines the probability paths as an optimal transport interpolation to straighten the trajectories, which leads to a similar formulation to the one used by Rectified Flow. Variational Rectified Flow Matching \cite{variational_rf_matching} was also proposed to improve trajectory straightness by explicitly modeling the multiple possible paths that cross a certain point using a VAE. A recent method QAC \cite{qac} conditions the flow model on learnable representation in order to reduce the trajectory curvatures. Despite the improvements these methods achieve, they still assume an independent coupling between the source and target distribution. In this work, we aim to reduce the curvatures by optimizing this coupling.

\paragraph{Optimized Forward Coupling.} Most related to our method are the works that explore the impact of the forward coupling on trajectory curvature. Some methods improve the coupling through approximating the optimal transport plan between the source and target distributions \cite{minibatch-ot, multisample-fm}. However, it is computationally infeasible to solve an optimal transport problem on an entire dataset, so they approximate on the mini-batch level. Fast-ODE \cite{fast-ode} explores parameterizing the coupling as a neural network as a function of the data sample and optimizing it jointly with the flow model, which is shown to minimize the forward intersections and hence leads to faster sampling. However, the unavailability of the data samples at in inference time restricts the representation of the learned coupling, since it cannot deviate significantly from the independent coupling. We propose a general formulation that subsumes Fast-ODE and allows for larger deviation from the independent coupling assumption.



\begin{figure*}[t!]
    \centering
    \includegraphics[width=0.9\textwidth]{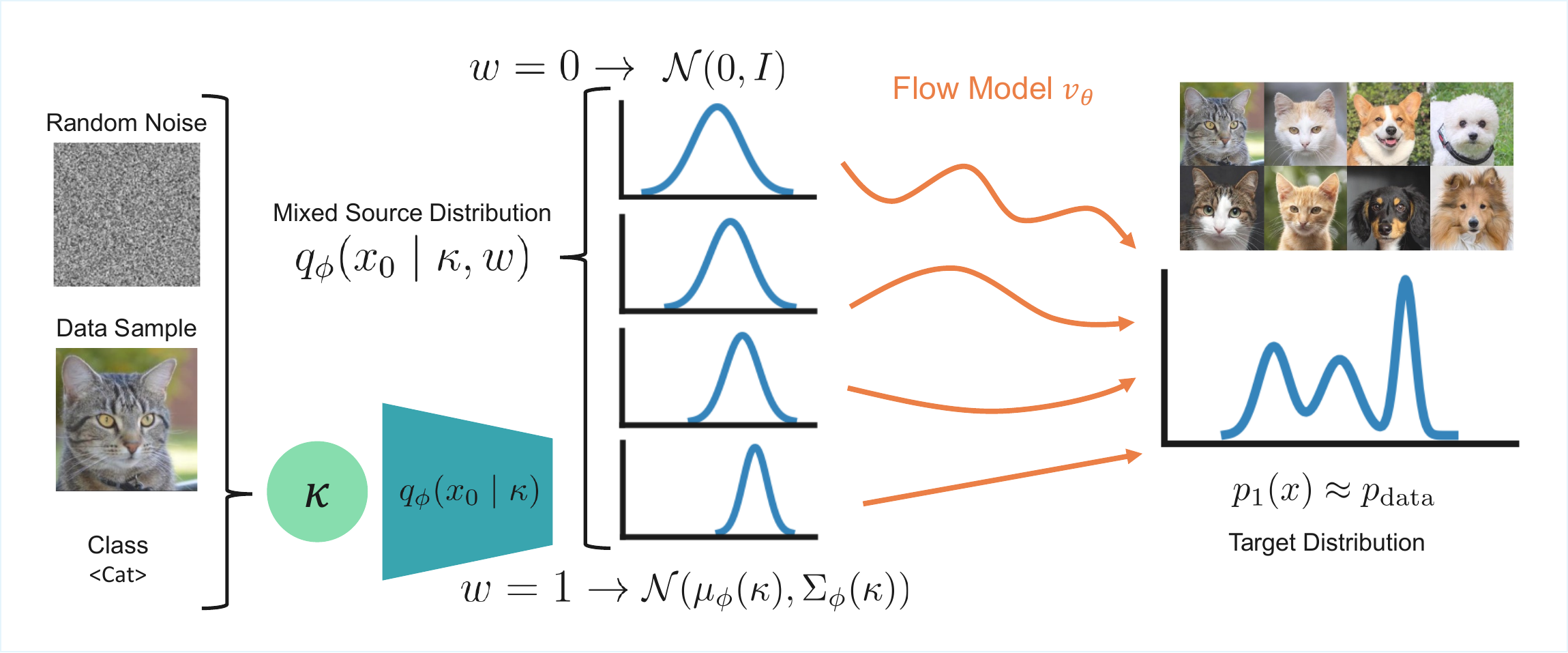}
    \vspace{-0.5cm}
    
    \caption{\textbf{Method overview.} We propose training rectified flows with mixed source distributions, obtained by interpolating a conditional and a simple unconditional distribution. The conditional distribution is predicted from a signal $\kappa$, which is possibly informative, e.g. a specific data example, a class label, or entirely independent, e.g. random noise. The learned conditional distribution provides a trajectory structure that minimizes the degree of intersections, which is inherited by the mapping from the unconditional source distribution.}
    \label{fig:method}
    \vspace{-0.3cm}
\end{figure*}
\section{Background}

    We first introduce necessary background and notations of Rectified Flows in Sec.~\ref{sec:rectified_flow} and the degree of trajectory intersection with its impact on sampling speed in Sec.~\ref{sec:intersection}.
    
    \subsection{Rectified Flow}
    \label{sec:rectified_flow}
        We assume a $d$-dimensional space $\mathbb{R}^d$ where the data points lie. The aim of Rectified Flow \cite{rectified-flow} is to learn a mapping between samples of a tractable source distribution $p_0(x)$ and a complex target distribution $p_1(x)$. We define $q(x_0,x_1)$ as their joint coupling whose marginal preserves their respective densities, and is by default assumed to be an independent coupling $q(x_0,x_1) = p_0(x)p_1(x)$. Given samples $x_0 \sim p_0(x), x_1 \sim p_1(x)$, an intermediate representation $x_t \sim p_t$ on the straight path between $x_0,x_1$ is defined as $ x_t = tx_1 + (1-t)x_0$ for $t \in [0,1]$, which represents a time-differentiable forward coupling between $p_0(x),p_1(x)$. 
        Rectified Flow proposes to learn a vector field $v_\theta(x_t,t)$ parametrized by $\theta$, which approximates the velocity required to flow in straight paths from $x_0$ to $x_1$, passing through $x_t$, defined as the time derivative $dx_t = v_t(x)dt = (x_1 - x_0)dt$ of the intermediate representation.
        The parameters $\theta$ of the learned vector field are found by minimizing

        \begin{align}
            \label{rf_loss}
            \mathcal{L}_{\text{RF}}(\theta) &:= \mathbb{E}_{x_0,x_1 \sim q(x_0,x_1)} \left[l(x_0,x_1)\right] , \:\:\: l(x_0, x_1) := \int_0^1  \left\lVert x_1 - x_0 - v_\theta(x_t,t) \right\rVert^2 dt \textnormal{.}
        \end{align}

    \subsection{Degree of Intersection}
\label{sec:intersection}
        Previous works~\cite{fast-ode, block-flow} have shown the effect of choosing the forward coupling $q(x_0,x_1)$ on the curvature of the generative trajectories. When the paths constructed between pairs $(x_0,x_1)$ in the forward process are highly intersecting, the vector-field model learns to estimate the mean direction, which causes the generative paths to be highly curved.
        
        The optimal $\theta^*$ in Eq.~\ref{rf_loss} is achieved when $v_{\theta^*}(x_t,t) = \mathbb{E}_{x_t}\left[x_1 - x_0 \mid x_t\right]$ as an estimator for the mean-squared error. Assuming we obtain an optimal model, and given a forward coupling $q(x_0,x_1)$, the degree of intersection of the forward trajectories can be estimated as:

        \begin{equation}
            I(q) = \mathop{\mathlarger{\mathbb{E}}}_{\substack{x_0,x_1 \sim q(x_0,x_1)}} \left[ \int_0^1  \left\lVert x_1 - x_0 - v_{\theta^*}(x_t,t) \right\rVert^2 dt\right] \textnormal{,}
            \label{doi}
        \end{equation}

        which is minimized for the same values as Eq.~\ref{rf_loss}~\cite{fast-ode}. With a fixed independent coupling \mbox{$q(x_0,x_1)=p_0(x)p_1(x)$}, $I(q)$ remains fixed. Therefore, in order to straighten the generated trajectories and improve sampling speed, previous methods \cite{minibatch-ot, fast-ode, multisample-fm} attempt to optimize $q(x_0,x_1)$ in order to minimize $I(q)$.

\section{Straightened Trajectories via Distribution Mixing}
    \vspace{-0.1cm}

This section introduces \meth{}, a method for training rectified flow from mixtures of (un)conditional source distributions. Fig.~\ref{fig:method} provides an overview of our approach.
In Sec.~\ref{sec:cond_src_dist}, we propose \FC{}, a general formulation of learnable forward couplings that can depend on arbitrary variables to optimize source distributions for a lower degree of intersection.
Moreover, we discuss the limitations of a naively constructed conditional source distribution relying on a simple Gaussian assumption.
To this end, Sec.\ref{sec:interpolation} formally introduces \meth{} and highlights its effect in overcoming these limitations.

\subsection{Learnable Forward coupling (\FC{})}
    \label{sec:cond_src_dist}
    Let $\kappa$ be a generic random variable that lies in $\mathbb{R}^n$. It can be an informative signal related to the data distribution $p_1(x)$, such as a class label, or entirely independent.
    In practice, $\kappa$ can represent  class labels, captions of an image, or any correlated or uncorrelated signal. Our general formulation subsumes the parametrization in Fast-ODE~\cite{fast-ode} as a special case, where the conditioning is the data sample itself $\kappa = x_1$. 

    Abstracting from the choice of $\kappa$, we assume it to be a common cause for $x_0$ and $x_1$, i.e., $x_0,x_1$ are conditionally independent given $\kappa$.
    With this assumption, the forward coupling can be written as:

    \begin{equation}
        \begin{split}
            q(x_0,x_1) = \int q(x_0,x_1 \mid \kappa)\:p(\kappa) d\kappa &= \int q(x_0 \mid \kappa) q(x_1 \mid \kappa) q(\kappa) d\kappa = \int q(x_0|\kappa)q(x_1,\kappa) d\kappa
        \end{split}
    \end{equation}

    Given this factorization, we propose a learnable coupling with additional parameters $\phi$: $q_\phi(x_0,x_1) = \int q_\phi(x_0 \mid \kappa) q(x_1,\kappa) d\kappa$.
    that can be jointly optimized with the vector field $v_\theta(x_t,t)$ to minimize the following loss

    \begin{equation}
        \mathcal{L}_{\text{\FC{}}}(\theta,\phi) = \mathbb{E}_{x_1,\kappa \sim q(x_1,\kappa),x_0 \sim q_\phi(x_0 \mid \kappa)} l(x_0,x_1) \mathrm{,}
        \label{eq:loss_kappa}
    \end{equation}
    
    which we obtain by sampling $x_0\sim q_\phi(x_0 \mid \kappa)$ in Eq.~\eqref{rf_loss}.
    One open design choice is the construction of $q_\phi(x_0 \mid \kappa)$. 
    A straightforward option is a Gaussian $q_\phi(x_0 \mid \kappa) = \mathcal{N}(\mu_\phi(\kappa), \Sigma_\phi(\kappa))$ with learnable mean and covariance, enforced by adding a regularization term $\beta D_{KL}\left(q_\phi(x_0|\kappa) \Vert \mathcal{N}(0, I)\right)$ to Eq.~\eqref{eq:loss_kappa}.
    However, we argue and empirically show (see Tab. \ref{tab:abl_cond_beta}) that this design strongly depends on the choice of the regularization weight $\beta$ that controls how close the conditional distribution is to the standard Gaussian.
    The case of $\beta\rightarrow0$ results in the prior hole problem~\cite{cvae}, well-known in the context of VAEs, that prevents sampling at inference time without a given $\kappa$.
    For $\beta\rightarrow\infty$, $q_\phi(x_0 \mid \kappa)$ becomes almost a standard Gaussian and therefore independent of $\kappa$ which removes any advantage of the learnable forward coupling.
    Even worse, $\beta$ is a hyperparameter determined prior training with strong influence on the performance during inference.
    
    Therefore, we propose an alternative design of $q_\phi(x_0 \mid \kappa)$ via distribution mixing that is robust to train and achieves lower curvature (cf. Sec.~\ref{sec:intersection}), resulting in higher quality samples with small numbers of network evaluations.






    
\subsection{Flowing from a Mixture of Two Distributions}
    \label{sec:interpolation}
    We propose to train the vector-field model to flow from linear interpolations of two distributions: (1) the parameterized Gaussian $\mathcal{N}(\mu_\phi(\kappa), \Sigma_\phi(\kappa))$ 
    , and (2) a standard Gaussian $\mathcal{N}(0, I)$. As linear interpolations of two Gaussians with scalar weight $w$, the resulting source distributions $q_\phi(x\mid \kappa,w) = \mathcal{N}(w\mu_\phi(\kappa),w\Sigma_\phi(\kappa) + (1-w)I)$ are themselves Gaussian.
    Training the vector field with a conditional source distribution, the unconditional standard normal, and everything in between enables the network to learn to effectively utilize $\kappa$ while enforcing full coverage of the Gaussian space. Thus, during training, the efficient coupling between conditional source distribution and target distribution is transferred to the unconditional source distribution as well, allowing inference without conditioning. We now introduce training and sampling algorithms.

\begin{wraptable}{r}{0.43\textwidth}
\vspace{-0.5cm}
  \centering
  \begin{tabular}{lcc}
    \toprule
    Model                                   & $\beta$           & Curvature ($\downarrow$) \\
    \midrule
     Rectified Flow \small   & $\infty$            &  0.0467 \\ 
    \midrule
    Fast-ODE \small               & 20                & 0.0388 \\
     MixFlow (ours)                         & $10^{-5}$         & \textbf{0.0366} \\
    \bottomrule
  \end{tabular}
    \caption{\textbf{Trajectory curvature.} We compare the generative trajectory of MixFlow with Rectified Flow and Fast-ODE. With lower $\beta$ coefficient for the KL divergence loss, MixFlow achieves $\sim 5\%$ improved curvature over Fast-ODE.}
  \label{tab:curvature}
  \vspace{-1.5cm}
\end{wraptable}
\begin{center}            
\begin{minipage}[t]{0.46\linewidth}
  \vspace{0pt}            
  \vspace{-0.4cm}
  \begin{algorithm}[H]
  \caption{Training}\label{alg:training}
  \DontPrintSemicolon
  \SetAlgoLined
  \SetKwInOut{Input}{Input}
  \SetKwInOut{Output}{Return}
  
  \Input{$q(x_1, \kappa), \mu_\phi, \Sigma_\phi, v_\theta, \beta, N$}

  \For{$i \leftarrow 1$ \KwTo $N$}{
    $x_1, \kappa \sim q(x_1,\kappa)$\, $t, w \sim \mathcal{U}(0,1)$\;
    $\mu_\kappa \gets \mu_\phi(\kappa), \Sigma_\kappa \gets \Sigma_\phi(\kappa)$\;
    $\mu_w \gets w\mu_\kappa$ , $\: \Sigma_w \gets w\Sigma_\kappa + (1-w)I$\;
    $x_0 \sim \mathcal{N}(\mu_w,\Sigma_w)$\;
    $x_t \gets tx_1 + (1-t)x_0$\;
    $\mathcal{L} \gets ||x_1 - x_0 - v_\theta(x_t,t)||^2 + \beta D_{KL}\left(\mathcal{N}(\mu_\kappa, \Sigma_\kappa) \Vert \mathcal{N}(0, I)\right)$\;
    Update ($\theta, \phi$) with $\nabla \mathcal{L}$\;
  }
  \Output{$\mu_{\phi^*}$, $\Sigma_{\phi^*}$, $v_{\theta^*}$}
  \end{algorithm}
\end{minipage}
\hfill
\begin{minipage}[t]{0.46\linewidth}
  \vspace{0pt}            
  \vspace{-0.4cm}
  \begin{algorithm}[H]
  \caption{Sampling}\label{alg:inference}
  \DontPrintSemicolon
  \SetAlgoLined
  \SetKwInOut{Input}{Input}
  \SetKwInOut{Output}{Return}

  \Input{$\mu_\phi$, $\Sigma_\phi$, $v_\theta$, $w$, $\kappa~\small{\text{(optional)}}$, \texttt{ODESolver}}

  \eIf{$\kappa$ is given}{
    $\mu_\kappa \gets \mu_\phi(\kappa), \Sigma_\kappa \gets \Sigma_\phi(\kappa)$\;
    $\mu_w \gets w\mu_\kappa$ , $\: \Sigma_w \gets w\Sigma_\kappa + (1-w)I$\;
    $x_{\text{init}} \sim \mathcal{N}(\mu_w, \Sigma_w)$
  }{
    $x_{\text{init}} \sim \mathcal{N}(0,I)$ 
  }\;
  $x_{\text{sampled}} \gets \texttt{ODESolver}(x_{\text{init}}, v_\theta)$ \;
  \Output{$x_{\text{sampled}}$}
  \end{algorithm}
\end{minipage}
\end{center}
    
    \paragraph{Training.} During training as outlined in Alg.~\ref{alg:training}, we sample $w \sim \mathcal{U}(0,1)$ independently for each example such that $v_\theta(x_t,t)$ learns to flow from a mixture of distributions, by minimizing

    \begin{equation}
        \mathcal{L}_{\text{ours}}(\theta,\phi) = \mathop{\mathlarger{\mathbb{E}}}_{\substack{
        x_1, \kappa \sim q(x_1,\kappa) \\ w\sim\mathcal{U}(0,1) \\
        x_0 \sim q_\phi(x_0 \mid \kappa, w)}} 
        \Big[l(x_0, x_1) + \beta R\left(x_0, \kappa, w\right)\Big]
        \mathrm{.}\label{loss_w}
    \end{equation}

    where $R\left(x_0, \kappa, w\right) = D_{KL}\left( q_\phi(x_0 \mid \kappa, w) \Vert \mathcal{N}(0, I)\right)$ is the KL divergence loss. This design has some important benefits. In Sec.~\ref{sec:abl_beta}, we show empirically that with this formulation, we can choose the regularization weight $\beta$ in Eq.~\eqref{loss_w} to be very small (order of $10^{-5}$) without losing training stability and coverage of the Gaussian prior for sampling. This essentially allows for a larger deviation of $q_\phi$ from the standard Gaussian distribution and therefore more complexity in its coverage of the data distribution modes. More importantly, the ability to use lower $\beta$ values allows for obtaining lower trajectory curvatures. See Tab. \ref{tab:curvature} and Supplementary A.2 for discussion and empirical evidence.
    
    \paragraph{Sampling.} For sampling as detailed in Alg.~\ref{alg:inference}, the source distribution can be chosen to be a Gaussian interpolant like during training, if $\kappa$ is available, or the standard normal as fallback. Furthermore, with a given $\kappa$, we can freely choose $w$ for sampling the initialization of the ODE. In Sec.~\ref{sec:abl_cond_src}, we show that $w$ provides control over the speed-quality tradeoff at inference time, which eliminates the need to retrain in order to push the performance with both low and high sampling budget. Even without $\kappa$ at inference time, e.g., in the case of $\kappa = x_1$ being the data sample itself, our method still achieves straightened sampling trajectories starting from the standard Gaussian source distribution.

\section{Experiments}
Our experimental evaluation of \meth{} comprises multiple established benchmark datasets for unconditional image generation. We further consider both low and high sampling budget settings using different ODE solvers to assess the achieved trade-offs between sampling quality and speed.

\subsection{Unconditional Generation on CIFAR10}

\begin{wraptable}{r}{0.5\textwidth}
\vspace{-0.4cm}
  \centering
  \begin{tabular}{lcccc}
    \toprule
    Method  & Solver & NFE & FID ($\downarrow$) \\ 
    \midrule
    \midrule
    Rectified Flow \small       & \multirow{6}{*}{RK45} & 127 & 2.58\\ 
    FM - OT \small     &  & 142 & 6.36\\ 
    Minibatch-OT \small            &  & 133.9 & 3.58\\ 
    Fast-ODE  \small      &  & 118 & 2.45\\ 
    QAC \small                            &  &  -  & 2.43\\ 
    \rowcolor{myrowcolour}
    \textbf{Ours}                       &  & 124.7 & \textbf{2.27}\\ 
    \midrule
    \midrule
    Fast-ODE  \small  & \multirowcell{2}{Heun's\\ $2^{nd}$ order}  & \multirow{3}{*}{5} & 24.40 \\ 
    QAC \small                             &  &  & 19.68 \\ 
    \rowcolor{myrowcolour}
    \textbf{Ours}                    &  &  & \textbf{19.29}\\ 
    \midrule
    \midrule
    Fast-ODE \small     & \multirowcell{2}{Heun's\\ $2^{nd}$ order} & \multirow{3}{*}{9} & 9.96     \\ 
    QAC \small                              &  &  & 10.28     \\ 
    \rowcolor{myrowcolour}
    \textbf{Ours}       &  &  & \textbf{8.97}    \\ 
    \bottomrule
  \end{tabular}
  \vspace{0.2cm}
    \caption{\textbf{Speed-quality trade-off on CIFAR10.} We evaluate our method on CIFAR10 in terms of FID using different ODE solvers. In the top, we fully simulate the ODE trajectory using the RK45 adaptive solver. \meth{} achieves notable improvements in FID with a comparable number of function evaluations (NFEs). Our method further improves sample quality with a small number of NFE as can be seen for 5 and 9 function evaluations with Heun's $2^{nd}$ order solver. 
    \vspace{-0.6cm}}  
  \label{tab:cifar10-main}
\end{wraptable}

     We demonstrate the effectiveness of our approach by comparing with previous methods on the CIFAR10 dataset~\cite{cifar10}. We train a rectified flow model with distribution mixing, where we define $\kappa$ as the data sample itself and choose $\beta = 10^{-5}$ in the loss \eqref{loss_w}. We follow the exact configuration of Fast-ODE \cite{fast-ode} for fair comparison.

    \subsubsection{Curvature Evaluation}
        We first evaluate the curvatures of the generative trajectories induced by the model trained with MixFlow and compare it with that of Fast-ODE and Rectified Flow. We generate 10K trajectories using an Euler sampler with 128 inference steps. Further details on the curvature computation can be found in the supplementary. We show the results in Tab. \ref{tab:curvature}. \meth{} achieves $\sim 22\%$ improvement compared to Rectified Flow and $\sim 5\%$ improvement compared to Fast-ODE. Due to our mixture formulation, we are able to train the vector-field model using a learned source distribution $q_\phi(x_0 \mid \kappa)$ with a much lower KL divergence weight ($\beta$), and hence it can deviate further from the standard Gaussian distribution, thereby achieving lower curvature in its generative paths.

    \subsubsection{Generation Evaluation}
    The evaluation w.r.t. Fréchet Inception Distance (FID) score covers both low and high sampling budget. We sample via full ODE simulation using the RK45 adaptive step-size solver for the high sampling budget. For the low number of sampling steps, we use the Heun $2^{nd}$ order solver to generate samples using 5 and 9 number of function evaluations (NFEs). The results are shown in Tab.~\ref{tab:cifar10-main}.

    \paragraph{Full Simulation.} We compare with trajectory curvature minimizing methods: Rectified Flow \cite{rectified-flow}, flow matching \cite{flow-matching}, QAC \cite{qac}, and also with methods that optimize the forward coupling:\cite{minibatch-ot}, Fast-ODE \cite{fast-ode}. Our method shows a reduction of $\sim 12\%$in FID compared to the standard Rectified Flow model, and $\sim 7\%$ improvement compared to the Fast-ODE with a comparable NFE. This shows that our method can better capture the diversity of the data distribution compared to previous methods.

    \begin{table}
  \centering
  \begin{tabular}{c|lcccccccc}
    \toprule
     & Model / NFE & $\beta$  & 4   & 5  & 10   & 20  & 32  & 64 & 128 \\
    \midrule
    \multirow{ 3}{*}{FFHQ} & \multirow{3}{*}{Fast-ODE }& 10 & \textbf{32.58} & \underline{25.33} & 13.21 & 8.85 & 7.54 & 6.91 & 7.01 \\
    &  & 20 & 38.23 & 29.12 & \underline{14.03} & 8.78 & 7.08 & 5.95 & 5.72 \\
    &  & 30 & 41.16 & 30.75 & 14.37 & \underline{8.76} & \underline{6.90} & \underline{5.45} & \underline{4.93} \\
    \midrule
    \rowcolor{myrowcolour}
    & \textbf{Ours} & $5 \times 10^{-5}$ & \underline{33.72} & \textbf{25.04} & \textbf{12.23} & \textbf{7.52 }& \textbf{5.31} & \textbf{4.01} & \textbf{3.75} \\
    \bottomrule
  \end{tabular}
  

    \begin{tabular}{c|lcccccccc}
    \toprule
    &Model / NFE & $\beta$ & 4   & 5  & 10   & 20  & 32  & 64 & 128 \\
    \midrule
    \multirow{ 3}{*}{AFHQ} & \multirow{3}{*}{Fast-ODE } & 10 & \underline{21.80} & \underline{18.04} & 11.80 & 9.05 & 8.22 & 7.47 & 7.21 \\
    & & 20 & 25.73 & 20.11 & \underline{10.56} & 6.89 & 5.74 & 4.92 & 4.55 \\
    & & 30 & 30.84 & 23.08 & 11.17 & \underline{6.66} & \underline{5.37} & \underline{4.40} & \underline{3.96} \\
    \midrule
    \rowcolor{myrowcolour}
    & \textbf{Ours} & $5 \times 10 ^{-5}$ & \textbf{19.72} & \textbf{15.57} & \textbf{7.95} & \textbf{5.05} & \textbf{4.30} & \textbf{3.65} & \textbf{3.33} \\
    \bottomrule
  \end{tabular}
  \vspace{0.2cm}
    \caption{\textbf{Comparison with Fast-ODE.} using with different KL divergence weights $\beta$ w.r.t FID-10k on the FFHQ and AFHQv2 $64 \times 64$ datasets. Our model outperforms Fast-ODE for different $\beta$ choices on almost all NFEs. \meth{} provides overall the best trade-off between sampling speed and quality without the need to retrain with a different $\beta$ parameter.
  }
  \label{tab:ffhq}
\end{table}
    
    \paragraph{Low Sampling Budget.} We compare with the most related and recent baselines Fast-ODE~\cite{fast-ode} and QAC~\cite{qac}. When NFE $=5$, our method achieves $\sim 20\%$ improvement compared to Fast-ODE in FID, and $\sim 2\%$ improvement compared to QAC. For NFE $=9$, our method improves FID by $\sim 10\%$ compared to Fast-ODE and $\sim 12.7\%$ improvement in comparison with QAC. This highlights the effectiveness of our approach in the low sampling budget regime.

\begin{figure*}[t!]
    \centering
    \vspace{-0.2cm}
    \includegraphics[width=0.75\textwidth]{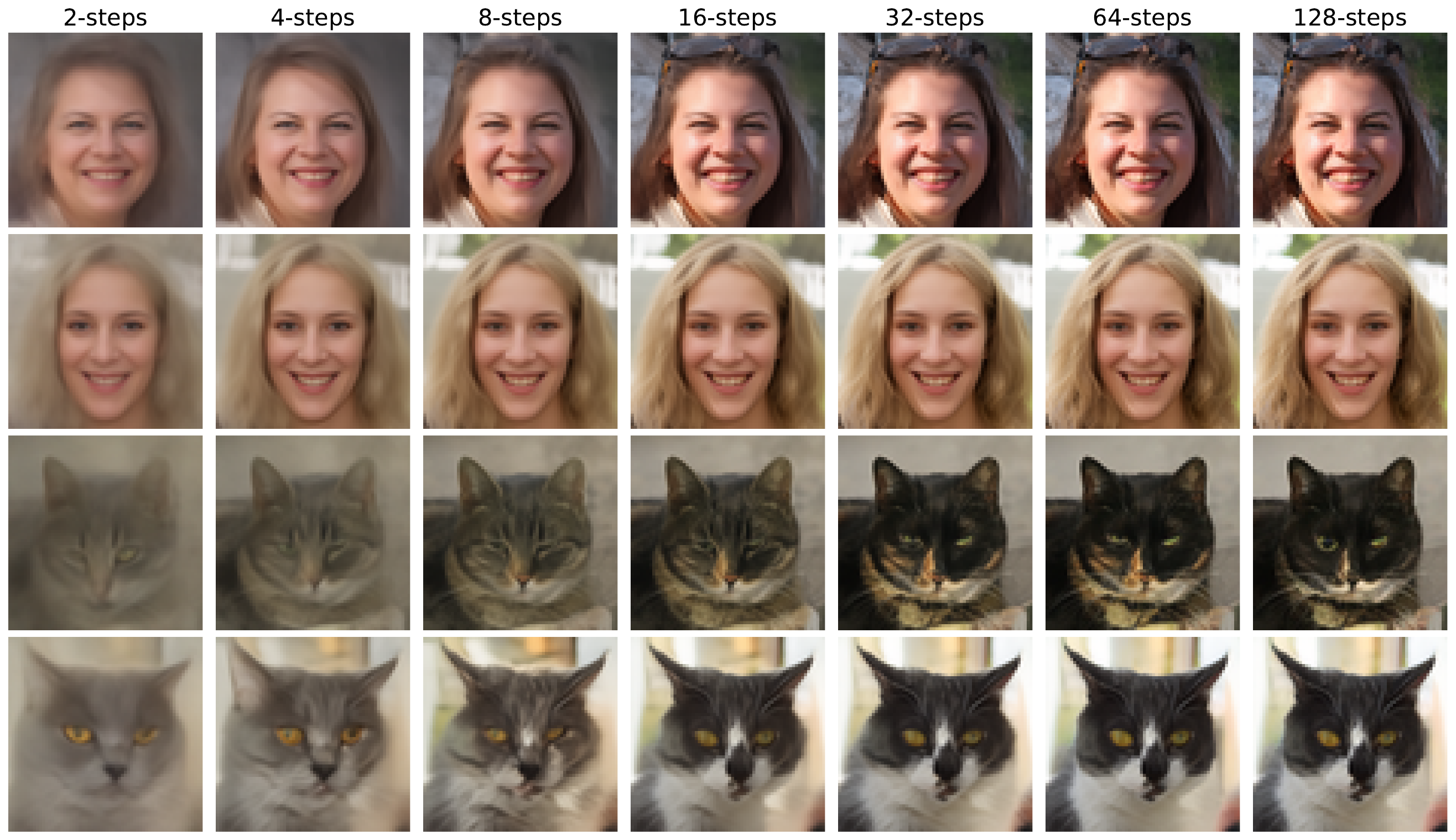}
    \caption{\textbf{Qualitative Results.} We show our method's generation on FFHQ (rows 1-2) and AFHQv2 (rows 3-4) datasets using different steps. \meth{} requires few steps to generate reasonable outputs.
    \vspace{-0.5cm}
    }
    \label{fig:qual}
\end{figure*}

\subsection{Unconditional Generation on FFHQ \& AFHQ}

    We further train and evaluate our method on higher resolution datasets FFHQ $64 \times 64$ \cite{ffhq_dataset} and AFHQv2 $64 \times 64$ \cite{afhq_dataset}. We train the models with $\kappa$ as the data sample, and with $\beta = 5 \times 10^{-5}$. We evaluate the performance with FID-10K using the Euler solver across different sampling steps. We compare against Fast-ODE \cite{fast-ode} which reports several results that differ in the choice $\beta$ in Tab. \ref{tab:ffhq}. In both datasets, \meth{} outperforms their models with different $\beta$ values for almost all NFEs. Despite the comparability of $\beta=10$ at low NFEs with \meth{}, its performance degrades with higher NFEs. On the other hand, our method achieves the best trade-off between speed and quality and improves the FID for all NFEs without the need to retrain with different $\beta$ values. In addition, we present qualitative examples for both datasets in Figure \ref{fig:qual}. With very few steps (<10), the generations are of already of a reasonable quality.

\section{Analysis}\label{sec:analysis}

\subsection{Choice of Conditioning for Source Distribution}
\label{sec:abl_cond_src}

    In addition to our default choice of $\kappa = x_1$ as the data sample itself, we explore two additional instantiations of $\kappa$. The first choice is defining $\kappa$ as the class label assigned to each sample of $p_1(x)$, we call it $\kappa_{c}$. While this choice seems to violate the assumptions of unconditional generation, our goal is to simply demonstrate the possibility of using a signal that is available during inference with our framework, which can motivate its effectiveness in more complex conditional generation tasks. Each class label is represented with a learnable embedding. The second choice explores the opposite of $\kappa = x_1$, where we assume $\kappa \sim \mathcal{N}(0,I)$ is a noise sample from a standard Gaussian distribution, referred to as $\kappa_n$, representing a case uncorrelated with the data distribution. 


\begin{table}[t]
\small
\centering

\begin{minipage}[t]{0.478\textwidth}
  \centering
  \resizebox{\textwidth}{!}{%
    \begin{tabular}{lccccccc}
      \toprule
      $\beta$ / \small{NFE}  & 2   & 4  & 10   & 20  & 32  & 64 & 128 \\
      \midrule
      $\infty$           & 171.7          & 54.5           & 13.16         & 6.90          & 5.02          & 3.63          & 3.04  \\
      $1$                & 168.35         & 53.64          & 13.29         & 6.86          & 4.97          & 3.58          & 3.02  \\
      $10^{-3}$          & 148.36         & 46.20          & 12.10         & 6.55          & 4.91          & 3.59          & 3.02 \\
      $10 ^{-5}$         & 99.30          & 29.64          & \textbf{9.02} & \textbf{5.23} & \textbf{3.90} & \textbf{2.95} & \textbf{2.52} \\
      $10^{-6}$          & 93.45          & 27.62          & 9.20          & 5.80          & 4.59          & 3.61          & 3.21 \\
      $5 \times 10^{-7}$ & \textbf{89.34} & \textbf{27.61} & 10.07         & 6.64          & 5.39          & 4.39          & 3.92 \\
      \bottomrule
    \end{tabular}%
  }
  \vspace{-0.2cm}
  \captionof{table}{\textbf{Effect of regularization weight $\beta$.} Lower $\beta$ values enabled by \meth{} result in improved FID. However, regularization is still required, as for smaller values in the order of $10^{-8}$, we observe that the source distribution collapses. We choose $\beta=10^{-5}$ as a default.}
  \label{tab:abl_cond_beta}
  \vspace{-0.2cm}
\end{minipage}
\hfill
\begin{minipage}[t]{0.478\textwidth}
  \centering
  \resizebox{\textwidth}{!}{%
    \begin{tabular}{lccccccc}
      \toprule
      Input / NFE  & 2   & 4  & 10   & 20  & 32  & 64 & 128 \\
      \midrule
      Rectified Flow       & 171.7  & 54.5  & 13.16 & 6.90 & 5.02 & 3.63 & 3.04 \\
      \midrule
      Noise $(\kappa_n)$   & 157.43 & 49.83 & 11.40 & 5.86 & 4.31 & 3.15 & 2.79 \\
      Label $(\kappa_c)$   & 160.17 & 48.65 & 11.35 & 5.89 & 4.37 & 3.27 & 2.82 \\
      Sample               & \textbf{99.30} & \textbf{29.64} & \textbf{9.02} & \textbf{5.23} & \textbf{3.90} & \textbf{2.95} & \textbf{2.52} \\
      \bottomrule
    \end{tabular}%
  }
  \captionof{table}{\textbf{Effect of conditioning signal $\kappa$.} $\kappa_n$ denotes $\kappa \sim \mathcal{N}(0,I)$ and $\kappa_c$ denotes the class label assumption. We sample from the $\kappa_n$ and $\kappa_c$ models with $w=0$. All choices of $\kappa$ lead to improvements over rectified flow (first row). Choosing $\kappa$ as the sample is best because it is maximally informative of the data distribution.}
  \label{tab:abl_cond_src}
  \vspace{-0.2cm}
\end{minipage}

\end{table}

    \paragraph{Effect on Performance.} We explore the impact of choosing $\kappa$ in comparison with the standard Rectified Flow model in Table \ref{tab:abl_cond_src}. For fair comparison, both models are evaluated with $w=0$, i.e,  with standard Gaussian. We see that the all choices of $\kappa$ improve FID compared to the baseline. Interestingly, we observe that an uninformative $\kappa_n$ (second row) can still improve FID for all sampling steps. This is due to the flexibility of the learnable forward coupling, which, through optimization in Eq.~\eqref{loss_w}, learns to map samples of $\mathcal{N}(0,I)$ to a sub-region that is more aligned with the data distribution. Even comparing $\kappa_n$ with $\kappa_c$, $\kappa_n$ slightly outperforms it. However, this advantage drops as $w$ increases (see supplementary). Nevertheless, defining $\kappa = x_1$ (last row) gives the best FID values across all steps, which reflects the importance of a maximally informed distribution. 

\subsection{Distance Between Mixed Distributions}
\label{sec:abl_beta}

        We explore the effect of the KL divergence weight hyperparameter $\beta$ in Eq.~\eqref{loss_w}. $\beta$ controls the deviation of the conditioned distribution from the standard Gaussian distribution. Therefore, it defines the width of the continuous range of that the model is exposed to. We train different models with different $\beta$ values on CIFAR10 and evaluate them using an Euler solver across several choices of sampling steps. Table~\ref{tab:abl_cond_beta} shows the results compared to the standard Rectified Flow model. We see that for all steps, the FID improves as $\beta$ decreases, achieving clear improvements at $\beta = 10^{-5}$. As $\beta$ goes below $10^{-5}$ the performance at low NFEs improves, while it deteriorates for high NFEs. 
        Hence, we find $\beta = 10^{-5}$ to provide considerable improvements while maintaining stability. Our results show that allowing the conditional distribution to deviate sufficiently from the standard Gaussian is essential to achieve faster sampling, which is uniquely enabled by our proposed \meth{}.


\section{Conclusion}

\label{sec:conclusion}
We addressed in this work the sampling efficiency problem of in flow models through the lens of curvature minimization. We presented \FC{}, a general formulation of learnable forward couplings for rectified flows that can leverage arbitrary signals. We highlighted the limitations by naively training with \FC{} and as a solution proposed \meth{} a training strategy that mixes conditional and unconditional distributions while training flow models. \meth{} successfully minimizes the trajectory curvature, improves performance under fixed sampling budget, and leads to faster convergence.

\paragraph{Limitations and Future Work.}
As our \FC{} formulation abstracts from a generic conditioning variable, we are excited to apply our method to other instances like text prompts, besides the current set of noise, labels, and data samples.
Furthermore, while \meth{} reduces the regularization of the learnable forward coupling (low KL divergence weight), which minimizes curvature and speeds up sampling, it still requires a Gaussian assumption. 
Therefore, we will investigate further relaxations while maintaining performance in future work.

\bibliography{iclr2026_delta}
\bibliographystyle{iclr2026_delta}

\appendix
\newpage
\appendix


\begin{center}
  {\LARGE\bfseries Supplementary Materials}
\end{center}
\vspace{1em}

\section{Trajectory Curvature}

    \subsection{Computation Details}
    
        \label{sec:curvature_details}
        In order to demonstrate the impact of our method MixFlow on the curvatures of the generative paths, we follow the procedure in Fast-ODE \cite{fast-ode} to estimate the curvature of the generated trajectories. We generate 10000 trajectories using an Euler solver with 128 steps, and compute the average curvature for an optimized vector-field model $v_\theta$ as:
    
        \begin{equation}
            C(v_\theta) = \mathbb{E}_{t, x_0} \left[ \left\lVert x_0 - x_1 - v_\theta(x_t,t) \right\rVert^2 \right]
        \end{equation}
    
        where $t \sim \mathcal{U}(0,1)$, $x_0 \sim p_0(x) = \mathcal{N}(0,I)$, and $x_1$ is sampled deterministically with an ODE solver: $x_1 = \texttt{ODESolver}(x_0, v_\theta)$.
        
        The curvature definition here closely resembles the degree of intersection $I(q)$ defined in eq. (4) in the main paper, where the differences are: (1) $I(q)$ is a function of the coupling $q(x_0,x_1)$ for the source and target distributions, whereas $C(v_\theta)$ is a function of an optimized vector field model,  (2): $x_1$ in $I(q)$ is sampled from the coupling, whereas $x_1$ in $C(v_\theta)$ is computed from $x_0$ deterministically with the ODE solver.

    \subsection{Effect of $\beta$}

\begin{figure}
        \centering
        \includegraphics[width=1.0\linewidth]{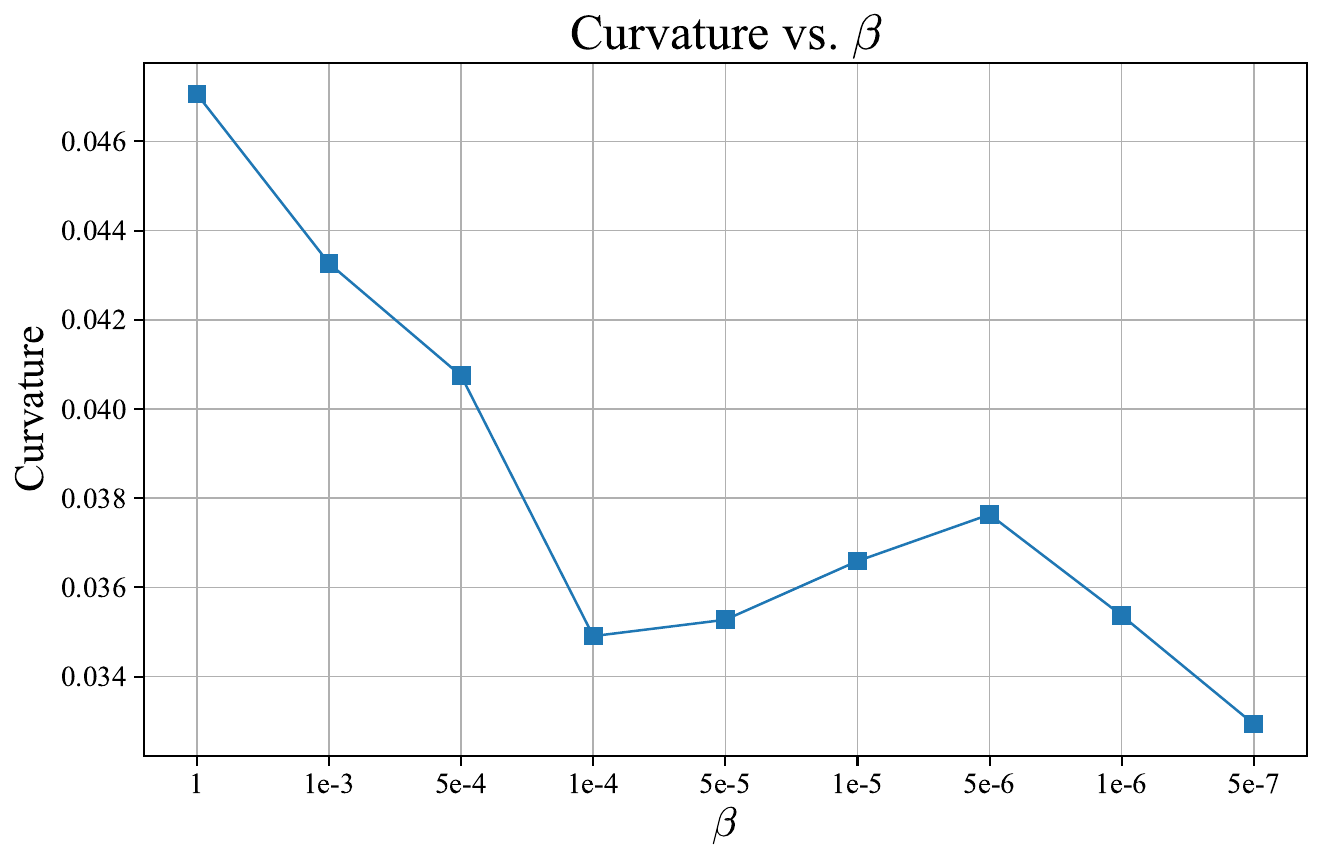}
        \caption{\textbf{Curvature vs. $\beta$.} We show how the curvature of the generative trajectories changes with different $\beta$ values. We can see a clear trend of lower curvature with a lower $\beta$ value}
        \label{fig:curvature_vs_beta}
    \end{figure}

        As we argue in the main paper, a main advantage of \meth{} is that it allows training our learnable forward coupling (\FC{}) with much lower KL Divergence weight $\beta$ compared to previous work, FastODE. Here, we show empirically that lower $\beta$ values correlate with lower curvatures of the generative trajectories. We train multiple \meth{} models with different $\beta$ values ranging from $5 \times 10ˆ{-7} $ to 1, and then evaluate the curvature of each model in order to see the trend. In Figure \ref{fig:curvature_vs_beta}, we see that as $\beta$ decreases (goes right on the x-axis), the curvature values (in the y-axis) tend to decrease.

\section{Implementation Details}
    
    \begin{table*}
  \centering
  \begin{tabular}{l|lccc}
    \toprule
       & Parameters                                              & CIFAR10 & FFHQ  & AFFHQv2   \\
    \midrule
    \multirow{7}{*}{UNet Parameters}    & Channel Size          & 128                   & 128                       & 128                           \\
                                        & Channel Multiplier    & [2,2,2]               & [1,2,2,2]                 & [1,2,2,2]                     \\
                                        & Blocks per Layer      & 4                     & 4                         & 4                             \\
                                        & Attention Resolution      & 16                    & 16                    & 16                            \\
                                        & Dropout Probability   & 0.13                  & 0.05                      & 0.25                          \\
                                        & Embedding type        & positional            & positional                & positional                    \\
                                        & Model Size            & 55.7M                 & 61.8M                     & 61.8                          \\
    \midrule
    \multirow{7}{*}{Training Setup}     & EMA ratio             & 0.9999                & 0.9999                    & 0.9999                        \\
                                        & Iterations            & 500K                  & 500K                      & 300K                          \\
                                        & Batch Size            & 128                   & 256                       & 256                           \\
                                        & Optimizer             & Adam      & Adam          & Adam              \\
                                        & Learning Rate (LR)    & $2 \times 10^{-4}$    & $2 \times 10^{-4}$        & $2 \times 10^{-4}$            \\
                                        & LR Scheduling         & constant              & constant                  & constant                      \\
                                        & LR Warmup steps       & 5000                  & 39060                     & 39060                         \\
    \bottomrule
  \end{tabular}
  \vspace{0.2cm}
    \caption{\textbf{Model and Experiments Configurations.} In the upper part, we show the UNet configuration for the vector field model $v_\theta$. In the lower part we show the training hyperparameters. Each column show the configuration for a specific dataset.   }
  \label{tab:implementation_config}
\end{table*}

    We share in this section the details about the models used in the experiments, as well as the training hyperparameters. 

    \paragraph{Vector Field Model $v_\theta(x_t,t)$.} We use the UNet architecture used in Fast-ODE \cite{fast-ode} for fair comparison, which follows the DDPM++ implementation of EDM \cite{edm}. You can see in Table \ref{tab:implementation_config}  the configuration of the network for used each dataset in the top part. In the bottom part, we show the corresponding training hyperparameters. All models optimized with Adam \cite{adam} with a learning rate that linearly increases until $2 \times 10^{-4}$ and then remains constant for the rest of the iterations. We also use Exponential Moving Average (EMA) on the model weights with ratio 0.9999, and we find it to be a critical factor for convergence. The CIFAR10 experiments all were trained on a single A100 80GB GPU. The FFHQ and AFHQv2 experiments each were trained on 4 A100 40GB GPUs, where the effective batch size shown in the tables was distributed equally among the GPUs.
    
    \begin{table*}
  \centering
  \begin{tabular}{lc}
    \toprule
    Parameters              &               \\
    \midrule
     Time Embedding Type    & positional    \\
     Flip Sin to Cos        & true          \\
     Down Block Types       & {\small [\texttt{DownBlock2D}, \texttt{DownBlock2D}, \texttt{DownBlock2D}, \texttt{AttnDownBlock2D}]} \\
     Up Block Types         & {\small [\texttt{AttnUpBlock2D}, \texttt{UpBlock2D}, \texttt{UpBlock2D}, \texttt{UpBlock2D}]} \\
     Block Out Channels     & [32, 64, 64, 64] \\
     Layers per Block       & 2                 \\
     Activation Function    & \texttt{silu}     \\
     Attention Head Dim     & 8                 \\
     Model Size             & $\sim 2$M  \\
    \bottomrule
  \end{tabular}
  \vspace{0.2cm}
    \caption{\textbf{Source Prediction Network $q_\phi(x_0 \mid \kappa)$ Configuration.} We show the UNet configurations for the source prediction parametrization. The params follow the \texttt{diffusers} library definition of the UNet model.
    }
  \label{tab:encoder_config}
\end{table*}

    \paragraph{Source Prediction Network $q_\phi(x_0 \mid \kappa)$.} We use a small UNet by adapting the \texttt{UNet2DModel} implementation from the \texttt{diffusers} library, version \texttt{0.32.2}. Its hyperparameters are shown in Table \ref{tab:encoder_config}, where the parameters are aligned with the library's implementation for ease of reproduceability. The input to the UNet is expected to be ($3 \times H \times W$), where $H,W$ can differ depending on the dataset used. However, depending on the input $\kappa$, the first layer might differ slightly. When $\kappa$ is the data sample (as in the default experiments) or a noise sample ($\kappa_n$), there is no change applied to the network.  When $\kappa$ is the class label ($\kappa_c$), an embedding layer is appended before, which maps the class labels into embeddings of size $3HW$, which are reshaped into $3 \times H \times W$ and then fed to the UNet. The network outputs the mean and the log variance of the distribution. We assume that the covariance is diagonal.

    \paragraph{Evaluation.} We use the Euler ODE solver from the \texttt{torchdiffeq} \cite{torchdiffeq} library, and its \texttt{scipy} \cite{scipy} library wrapper for the RK45 solver, where we set the \texttt{rtol} and \texttt{atol} parameters both to $10^{-5}$. As for Heun's $2^{nd}$ solver, we follow the manual implementation as in Fast-ODE \cite{fast-ode}. 

\begin{figure*}[ht]
  \centering
  \begin{subfigure}[b]{0.45\linewidth}
    \centering
    \includegraphics[width=\linewidth]{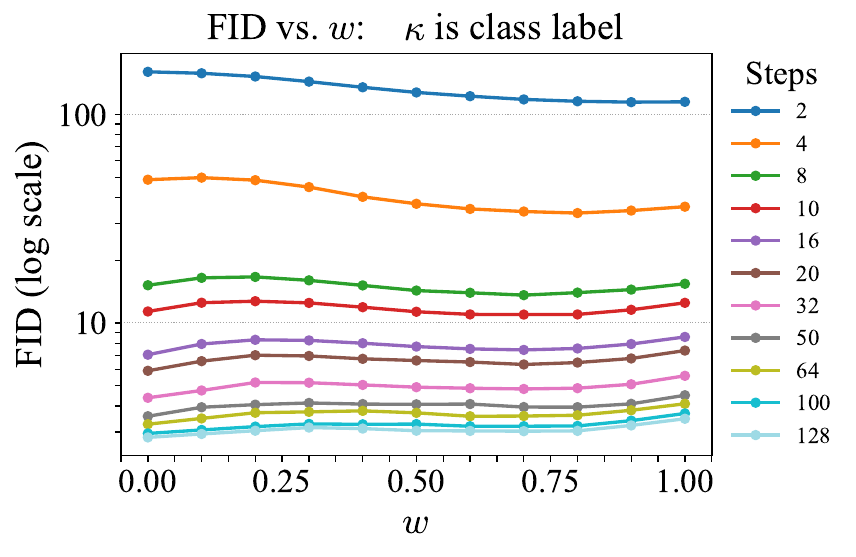}
    \vspace{-0.5cm}
    \caption{
    $\kappa_c$: Class label conditioning}\label{fig:fid_vs_w_label}
  \end{subfigure}
  \begin{subfigure}[b]{0.45\linewidth}
    \centering
    \includegraphics[width=\linewidth]{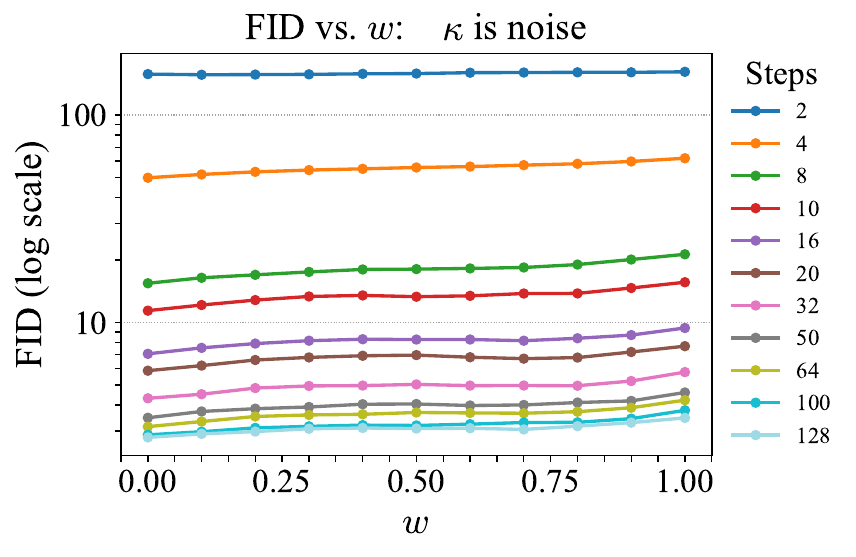}
    \vspace{-0.5cm}
    \caption{$\kappa_n$: Noise conditioning}\label{fig:fid_vs_w_noise}
  \end{subfigure}

  \caption{\textbf{Effect of interpolation weight $w$.} We visualize the effect of varying $w$ (x-axis) during sampling on FID (y-axis) across different numbers of sampling steps (different lines). \textbf{(a)} Sampling with a few steps benefits from a larger weight of the source distribution conditioned on class label $\kappa_c$, while for many steps, the unconditional standard Gaussian is better suited.  
  \textbf{(b)} With conditioning on uncorrelated Gaussian noise $\kappa_n$, the best FID is achieved for $w=0$, i.e., not using the conditional distribution at all during sampling. However, note that even in this case training the vector field on a mixture of distributions still improves performance as shown in Tab.~\ref{tab:abl_cond_src}.
  \vspace{-0.2cm}}
  \label{fig:fid_vs_w}
\end{figure*}

\begin{figure*}[ht]
  \centering
  \begin{subfigure}[b]{0.48\linewidth}
    \centering
    \includegraphics[width=\linewidth]{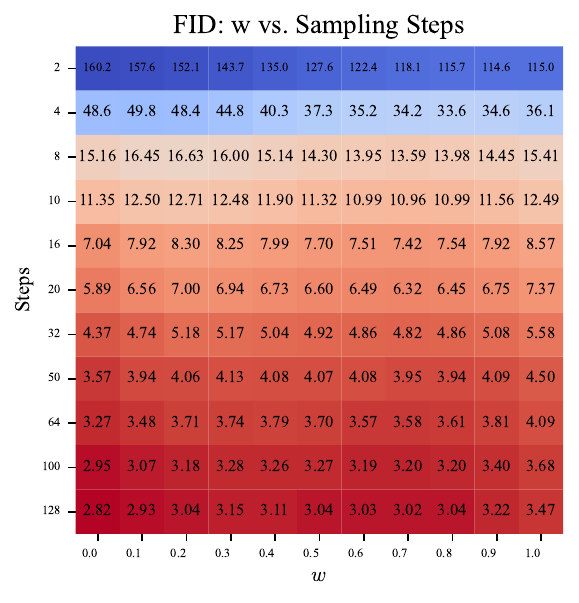}
    \vspace{-0.7cm}
    \caption{
    $\kappa_c$: Class label conditioning}\label{fig:fid_vs_w_label_hist}
  \end{subfigure}
  \begin{subfigure}[b]{0.48\linewidth}
    \centering
    \includegraphics[width=\linewidth]{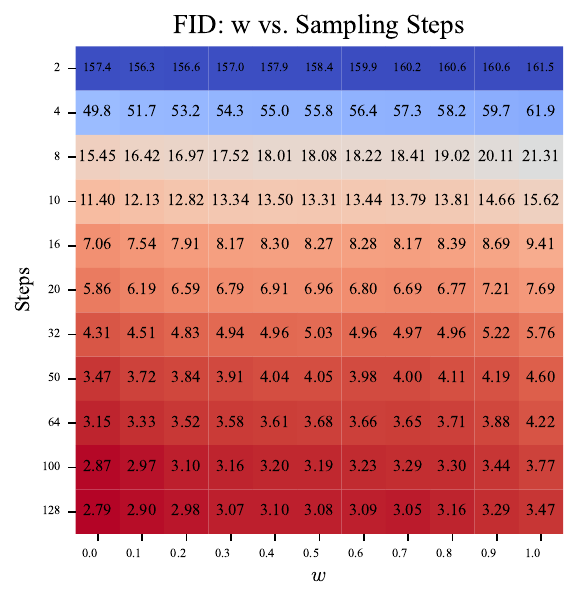}
    \vspace{-0.7cm}
    \caption{$\kappa_n$: Noise conditioning}\label{fig:fid_vs_w_noisel_hist}
  \end{subfigure}

  \caption{\textbf{FID for Sampling Steps vs. weight $w$.} We show the FID across different sampling step choices for both $\kappa_c$ and $\kappa_n$ as the interpolation parameter $w$ changes. This is the numerical version of Figure 3 in the main paper, which provides a more accurate look into the change of FID values, where red indicates lower FID and blue indicates higher FID.
  }
  \label{fig:fid_vs_w}
\end{figure*}

\begin{figure}
  \centering
 \vspace{-0.5cm}
    \includegraphics[width=\linewidth]{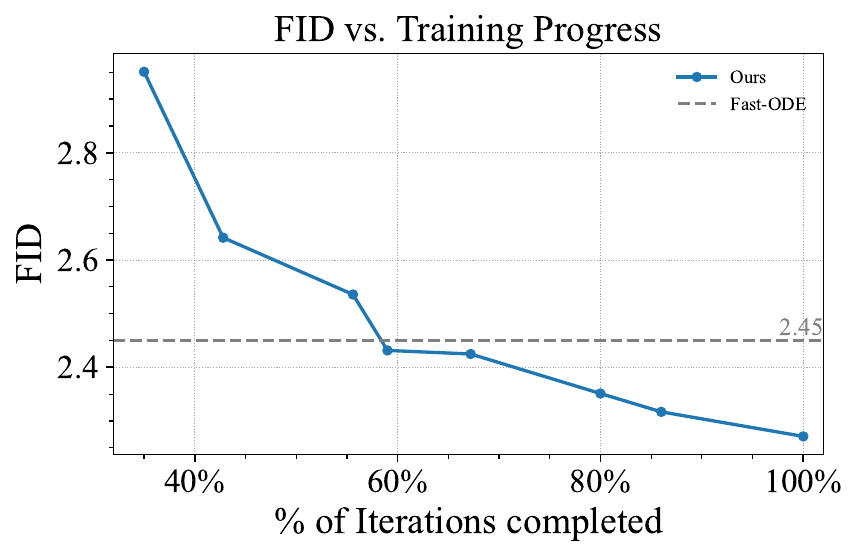}
    \caption{\textbf{FID vs. training progress.} Samples are generated with the RK45 solver across different step of the training process. Our method achieves the same performance as Fast-ODE (gray dotted line) with only 60\% of the training budget.
    \vspace{-0.4cm}
    }
\label{fig:fid_vs_iter}
\end{figure}

\section{Additional Analysis}

    \subsection{Training Efficiency}
    \label{sec:abl_iter}
    
        We highlight the training efficiency as another notable advantage of our formulation. Fig.~\ref{fig:fid_vs_iter} shows the FID during training compared to the final performance of one of our strongest baselines Fast-ODE~\cite{fast-ode}. Note that \meth{} achieves approximately the same performance as Fast-ODE using only $60\%$ of the full training iterations.
        Therefore, our method not only accelerates sampling by straightening flow trajectories, but also training in terms of convergence.

    \subsection{Effect of $w$}

        In the cases where the conditioning signals are available during inference, we can use them to vary the mixture parameter $w$ while sampling. So to understand the importance of choosing $w$, we train two different models with $\kappa_c,\kappa_n$. Figure \ref{fig:fid_vs_w_noise} shows the results for $\kappa_n$. We can see that the FID (y-axis) achieves the best value at $w=0$ and then degrades as $w$ (x-axis) increases from $0$ to $1$, showing that when the signal is not very informative, flowing from the standard Gaussian distribution where $w = 0$ results in better generation. As for $\kappa_c$, as shown in Figure \ref{fig:fid_vs_w_label}, an interesting pattern appears. For a low number of sampling steps (2,4), we notice that the FID tends to improve as $w$ increases, showing that the distribution learned from the informative signal has a positive effect with few sampling steps. As the number of sampling steps increases, we see that the FID is best at $w=0$ and starts dropping as $w$ increases. Hence, we conclude that, with a sufficiently informative signal, $w$ can control the quality-speed tradeoff during inference. So, depending on the sampling budget available, $w$ can be tuned at inference to provide the best FID accordingly.
        
        We also provide in Figure \ref{fig:fid_vs_w} a more fine-grained version of the figure to clearly see the difference in FID values between the two choices: $\kappa_c$: class label, and $\kappa_n$: standard Gaussian noise. The histograms show the FID values for the choice of sampling steps (y-axis) against the mixture parameter $w$ (x-axis). Stronger redness indicates lower FID, and stronger blueness indicates higher FID. When $w=0$,  we notice that $\kappa_n$ performs slightly better than $\kappa_c$, as shown in Table 3 in the main paper. As $w$ increases, however, the performance of $\kappa_c$ becomes better for all sampling steps compared to $\kappa_n$, which reflects the effect of class labels as a conditioning signal.

    \subsection{Sampling Steps}

\begin{figure*}[htbp]
  \centering
  
  \begin{subfigure}[b]{\textwidth}
    \centering
    \includegraphics[width=\textwidth]{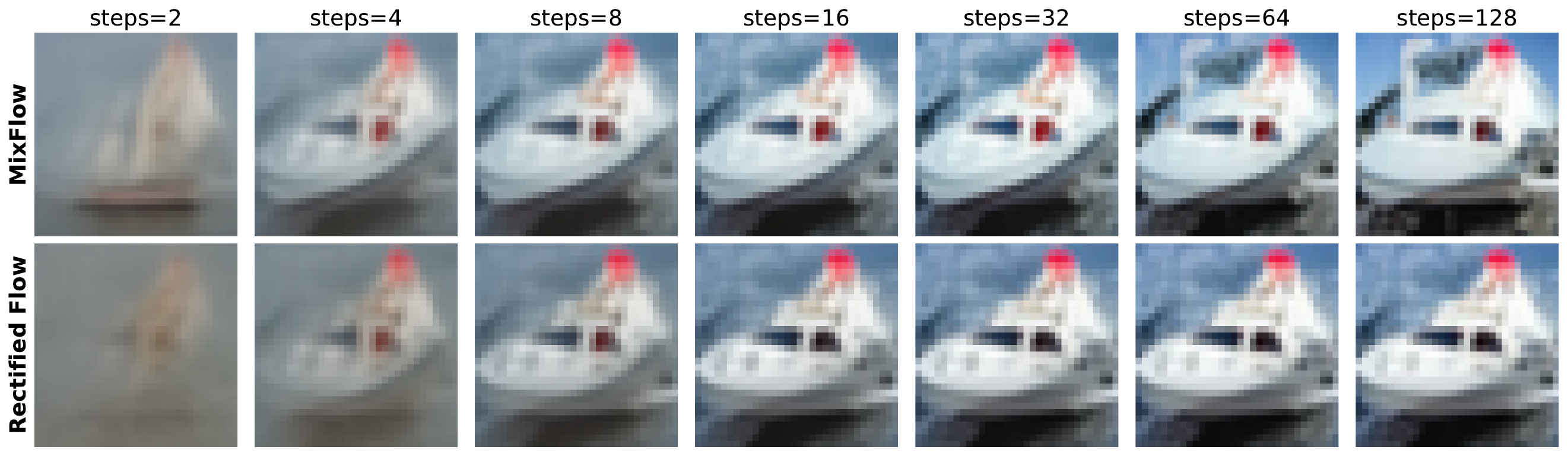} 
    \label{fig:sub1}
  \end{subfigure}
  
  \begin{subfigure}[b]{\textwidth}
    \centering
    \includegraphics[width=\textwidth]{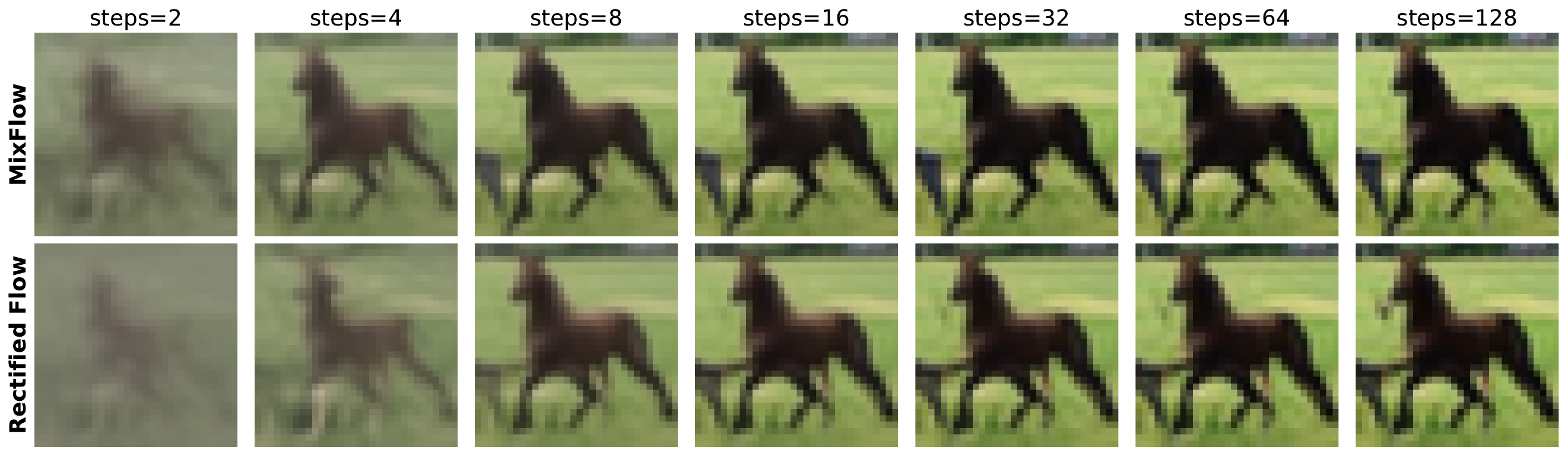}
    \label{fig:sub2}
  \end{subfigure}
  
  \begin{subfigure}[b]{\textwidth}
    \centering
    \includegraphics[width=\textwidth]{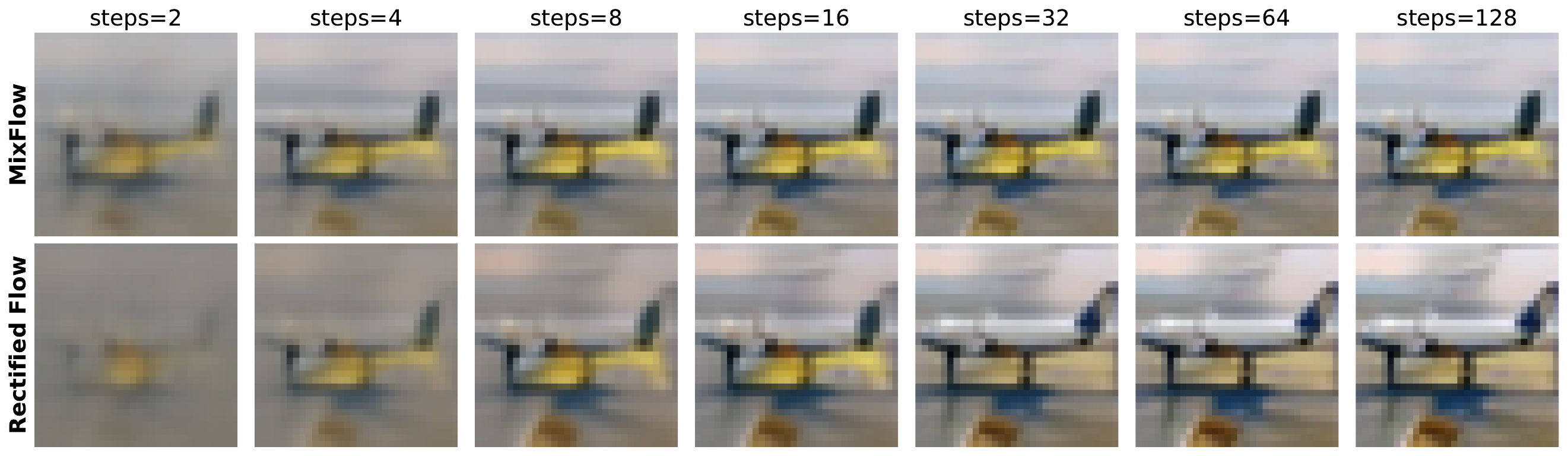}
    \label{fig:sub3}
  \end{subfigure}
  
  \caption{\textbf{Comparison against Rectified Flow}. We show multiple examples for generated images with different sampling steps and compare against Rectified Flow. We highlight that for a low number of sampling steps (2,4), MixFlow generates much clearer images compared to Rectified Flow.}
  \label{fig:ours_vs_rf}
\end{figure*}

        In order to highlight the effectiveness of MixFlow for improving sampling speed, we show in Figure \ref{fig:ours_vs_rf} some qualitative generations across different sampling step choice in comparison with Rectified Flow. Notice that with low sampling steps (2,4), MixFlow generates higher quality samples in comparison with those of Rectified Flow.

\section{Additional Qualitative Results}

\begin{figure*}[t!]
    \centering
    \includegraphics[width=\textwidth]{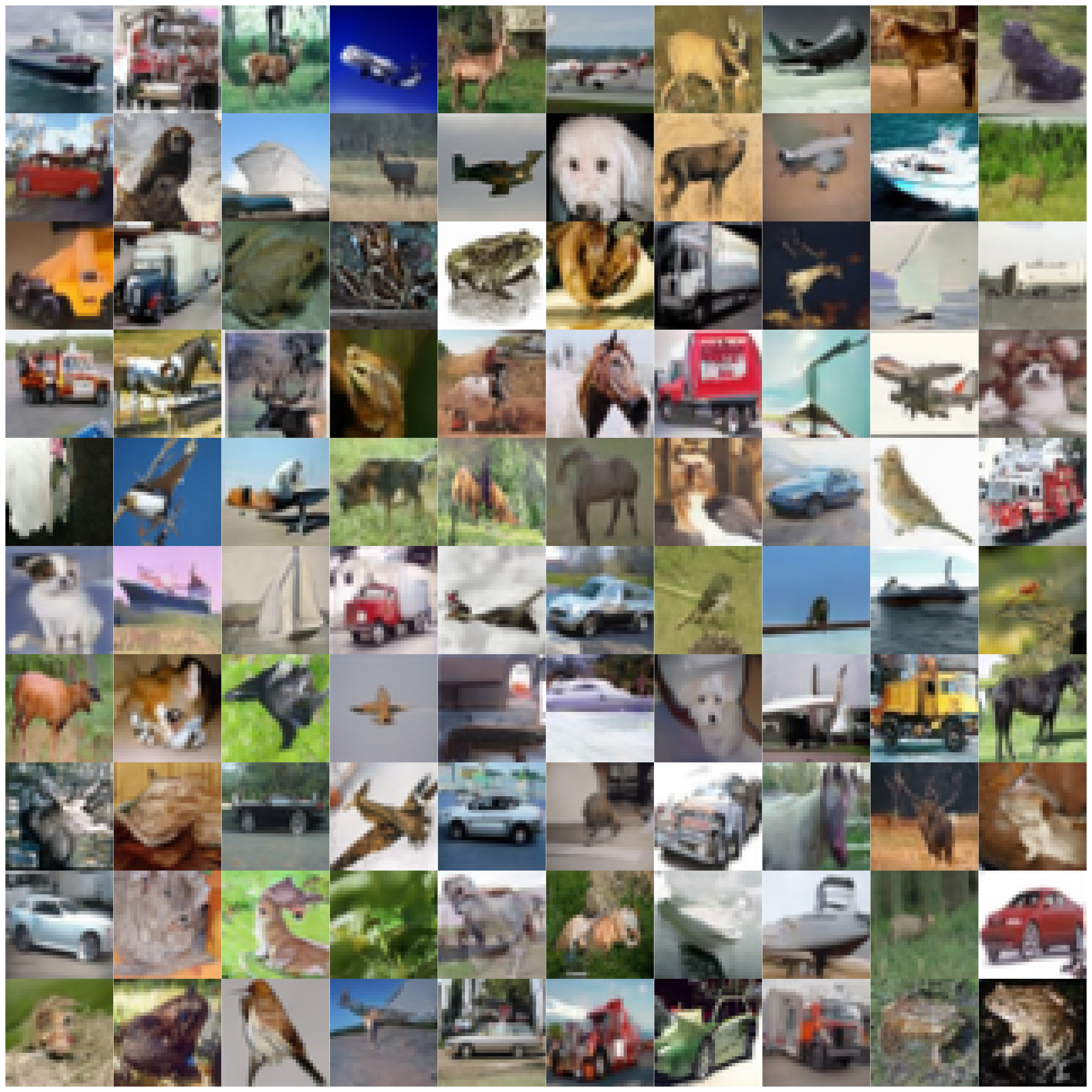}
    \vspace{-0.5cm}
    
    \caption{\textbf{Qualitative Results on CIFAR10}}
    \label{fig:cifar10_qual}
    \vspace{-0.3cm}
\end{figure*}

\begin{figure*}[t!]
    \centering
    \includegraphics[width=\textwidth]{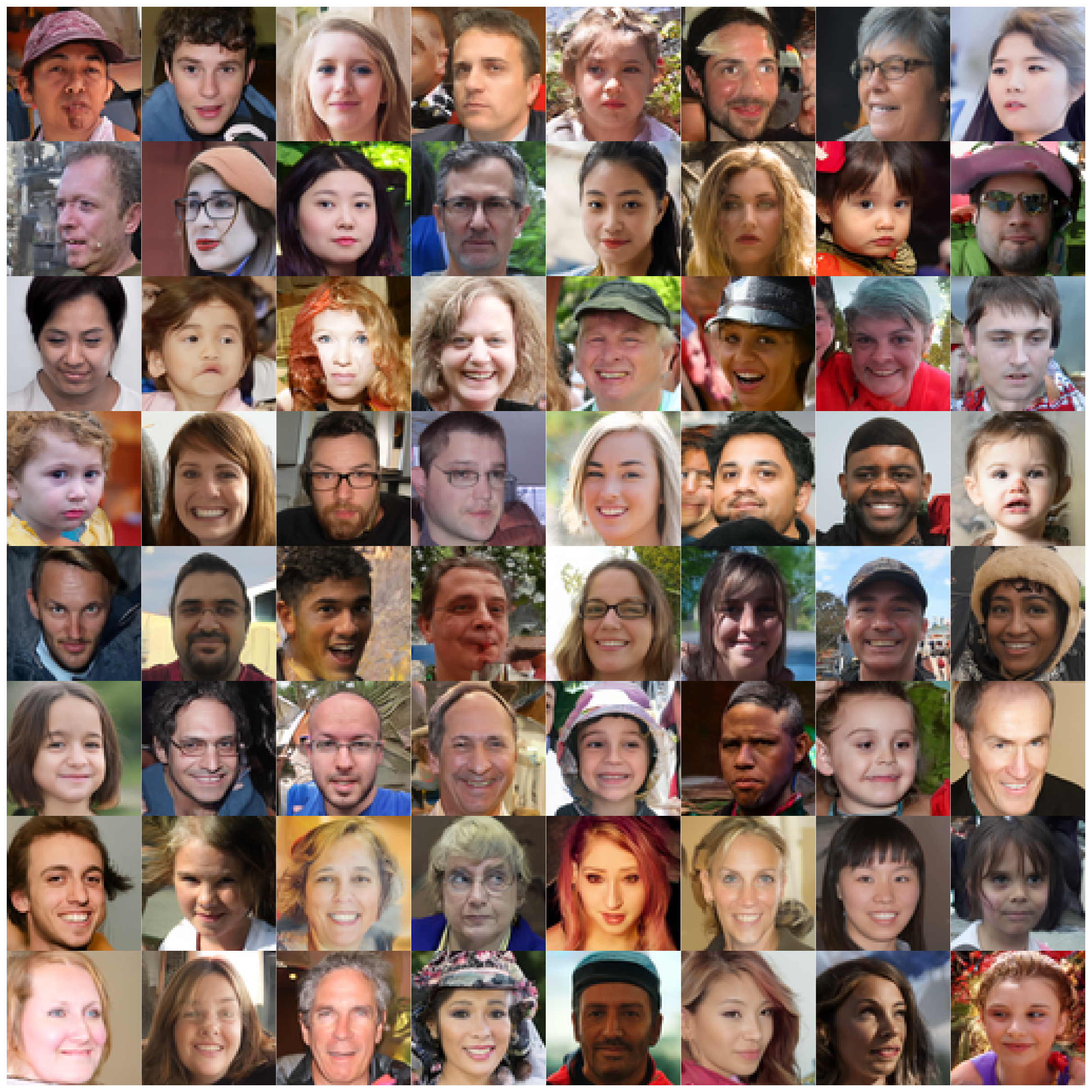}
    \vspace{-0.5cm}
    
    \caption{\textbf{Qualitative Results on FFHQ $64 \times 64$}}
    \label{fig:ffhq_qual}
    \vspace{-0.3cm}
\end{figure*}

\begin{figure*}[t!]
    \centering
    \includegraphics[width=\textwidth]{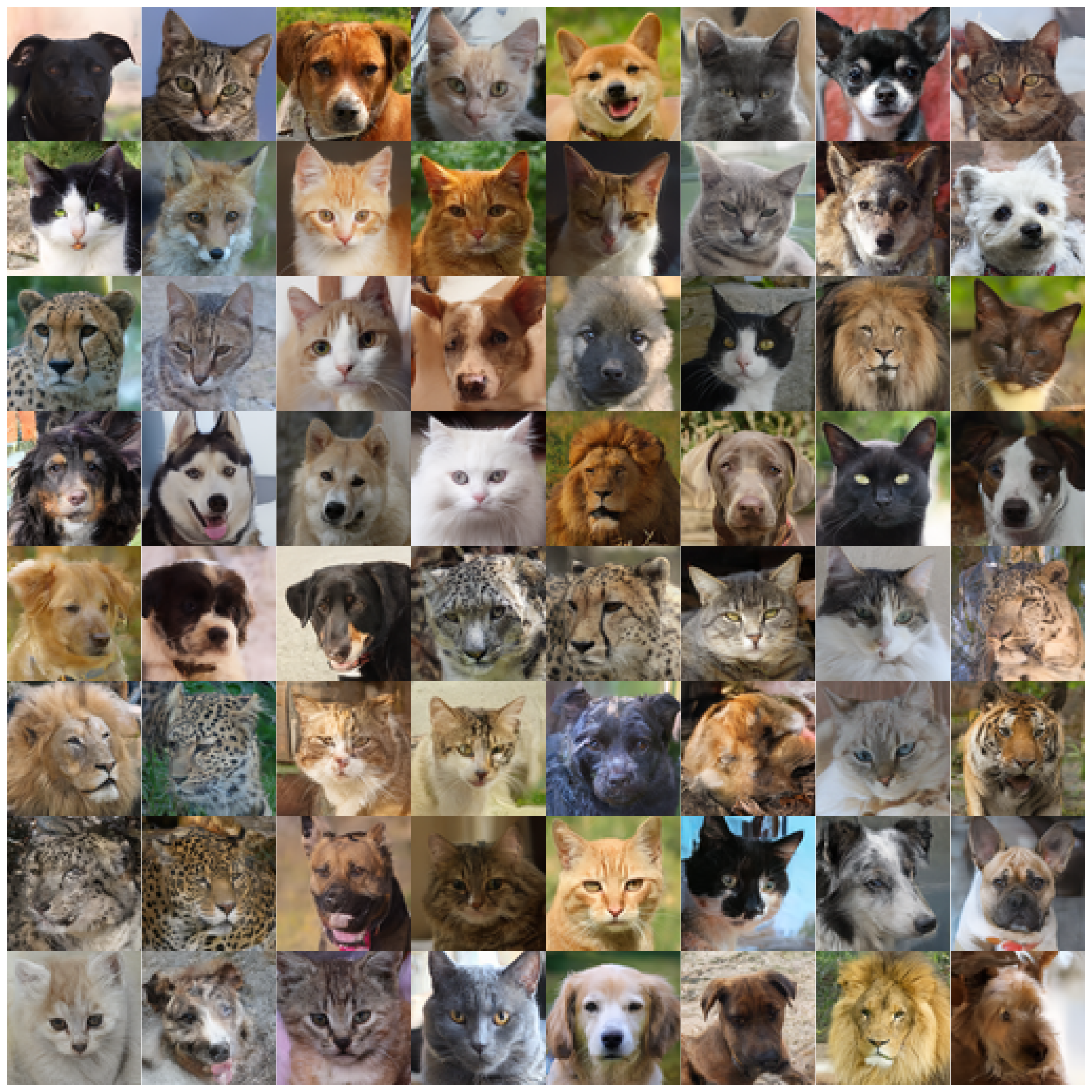}
    \vspace{-0.5cm}
    
    \caption{\textbf{Qualitative Results on AFHQv2 $64 \times 64$}}
    \label{fig:afhq_qual}
    \vspace{-0.3cm}
\end{figure*}

    We include more qualitative examples that are generated with MixFlow when $\kappa$ is the data sample, and $\beta=10^{-5}$. All the images are generated with Euler solver with 64 sampling steps. We show the generations for CIFAR10 in Figure \ref{fig:cifar10_qual}, FFHQ $64 \times 64$ in Figure \ref{fig:ffhq_qual}, and AFHQv2 $64 \times 64$ in Figure \ref{fig:afhq_qual}.

\end{document}